\documentclass[a4paper,twoside]{article}

\usepackage{epsfig}
\usepackage{graphicx}
\usepackage{subcaption}
\usepackage{calc}
\usepackage{amssymb}
\usepackage{amstext}
\usepackage{amsmath}
\usepackage{amsthm}
\usepackage{tabularx}
\usepackage{multirow}
\usepackage{multicol}
\usepackage{pslatex}
\usepackage{apalike}
\usepackage[bottom]{footmisc}
\usepackage{SCITEPRESS}     

\begin{document}

\title{A Comparison of Automatic Labelling Approaches for Sentiment Analysis}
\author{
	\authorname{Sumana Biswas\sup{1}, Karen Young\sup{1}and Josephine Griffith\sup{1}}
	\affiliation{\sup{1}School of Computer Science, National University of Ireland, Galway, Ireland }
	\email{\{s.biswas2,karen.young,josephine.griffith\}@nuigalway.ie}
}

\keywords{Sentiment Analysis, NLP, Deep Learning, Automatic Labelling}

\abstract{Labelling a large quantity of social media data for the task of supervised machine learning is not only time-consuming but also difficult and expensive. On the other hand, the accuracy of supervised machine learning models is strongly related to the quality of the labelled data on which they train, and automatic sentiment labelling techniques could reduce the time and cost of human labelling. We have compared three automatic sentiment labelling techniques: TextBlob, Vader, and Afinn to assign sentiments to tweets without any human assistance. We compare three scenarios: one uses training and testing datasets with existing ground truth labels; the second experiment uses automatic labels as training and testing datasets; and the third experiment uses three automatic labelling techniques to label the training dataset and uses the ground truth labels for testing. The experiments were evaluated on two Twitter datasets: SemEval-2013 (DS-1) and SemEval-2016 (DS-2). Results show that the Afinn labelling technique obtains the highest accuracy of 80.17\% (DS-1) and 80.05\% (DS-2) using a BiLSTM deep learning model. These findings imply that automatic text labelling could provide significant benefits, and suggest a feasible alternative to the time and cost of human labelling efforts.}

\onecolumn \maketitle \normalsize \setcounter{footnote}{0} \vfill
\section{\uppercase{Introduction}}
\label{sec:introduction}
Social media facilitates the sharing of ideas, views, and emotional responses between people. In these interactions, people often use short-text, meaningless unofficial words, short words, and emoticons, which can make it confusing to understand the exact meaning of the text. Sentiment analysis can be defined as a technique for classifying emotions into binary (Positive, and Negative) or ternary (Positive, Negative, and Neutral) classes. Extracting sentiment from this typically unstructured social media data is challenging and presents a difficult problem for researchers in this space. Analysing opinions is important because it can provide useful information for specific products, perspectives, or commentary on world events. Successful sentiment analysis of textual information \cite{daniel2017company,xu2019sentiment} depends on three main elements: a meaningful and clear expression of context; correctly labelled training and test datasets; and a suitable machine learning algorithm capable of accurately characterizing sentiment.

The labelling of social media data is an open problem for researchers when analysing sentiment. Although, it is easy and cheap to get labelled data from some online crowdsourcing systems like Rent-A-Coder, Amazon Mechanical Turk, etc. \cite{snow2008cheap,whitehill2009whose}, in many cases, one has no idea about the availability, efficiency, and quality of the labellers for a specific field or task. Labelling errors from non-expert labellers, and biased labelling due to a lack of efficiency, can result in incorrect and imbalanced labelling. The alternative, to ensure high-quality labels, is to engage human experts to perform the labelling, but this is both time-consuming and expensive. Furthermore, Sentiment analysis approaches are evaluated on datasets with automatic labels \cite{saad2021determining,chakraborty2020sentiment}.
However, there is a big gap in the evaluation of using automatic labelling versus human labelling to obtain acceptable predictions using machine learning algorithms. There has been very little study done to support or recommend the best labelling method. The lack of evaluation of these automatic labelling approaches motivates our work.

In this study, we have compared three state-of-the-art automatic labelling methods with the intention of quickly producing sentiment labels for Twitter data without human involvement. We have evaluated our approaches on two larger datasets from the SemEval-13 and SemEval-16 competitions, which contain ground truth and act as our human labelling. In this paper, we will use the term ‘Human Labelling'(HL) as ground truth throughout the discussion. The labels: ‘Positive’, ‘Negative’, and ‘Neutral’ are classified and assigned by the analysis of word meanings and polarity scores according to the word features and patterns. Three lexicon-based state-of-the-art automatic labelling approaches: TextBlob, Afinn, and Vader are used to generate all sentiment labels. The results of automatic labelling are compared to the results of human labelling using deep learning algorithms for sentiment analysis in order to determine reasonable alternatives to labelling social media data. The main objectives of our work are as follows:
\begin{itemize}
	\item Assign sentiment labels to the Twitter datasets using three automatic labelling methods: Afinn, TextBlob, and Vader.
	\item To establish whether automatic labelling approaches would be a viable alternative as compared with human labelling in terms of reducing the time and expense of human labelling.
	\item To obtain the best deep learning algorithm to analyse sentiment of Twitter datasets when the three automatic labellings are applied individually to the three state-of-the-art deep learning algorithms.
	
\end{itemize}  
The paper is organized as follows: in Section 2, related work on labelling strategies for state-of-the-art classifier methods and sentiment analysis approaches is presented. Section 3 describes our methodology, while Section 4 details the experiment and discusses the results. Finally, Section 5 concludes the paper with a discussion of limitations and plans for future work.
\section{\uppercase{Related Work}}
\label{sec:related work}
Sentiment analysis is a challenging problem due to the features of natural languages, such as the use of words in different situations, indicating different meanings. Sentiment analysis approaches are evaluated on datasets with either human labels \cite{mohammad2016semeval,mohammad2017stance,deriu2016swisscheese} or automatic labels \cite{saad2021determining,chakraborty2020sentiment}. Human annotators assign the sentiment labels for training and testing datasets by their understanding and expertise. Most of the datasets are annotated by human experts for relatively small datasets. For example, \cite{deriu2016swisscheese,mohammad2016semeval} used human labelling to predict sentiment. But, Oberlander and Klinger (2018) found noisy labels \cite{sheng2008get} in the largest  human annotated dataset (39k) \cite{web:lang:stats}. Several authors \cite{dimitrakakis2008cost,lindstrom2010handling,turney1994cost} studied data labelling costs and found that obtaining high-quality labelled data is time-consuming and cost-ineffective. The alternative of obtaining sentiment labels using automatic labelling saves time and money. The lexicon-based automatic labelling techniques of TextBlob, Vader, and Afinn use the NLTK (Natural Language Toolkit) of python libraries to automatically classify the sentiment of text and have been used in many studies. Each lexicon-based sentiment labelling approach needs a predefined word dictionary to infer the polarity of the sentiment according to their rules. Deepa et al. (2019) assessed the polarity scores of words to categorize the sentiment for the Twitter dataset related to UL Airlines with human labels using two dictionary-based methods: Vader dictionaries and Sentibank; the Vader dictionaries outperformed Sentibank by 3\% in their analysis to detect the polarity scores for the sentiment classification using the Logistic Regression (LR) model. TextBlob and Afinn were used by \cite{chakraborty2020sentiment} to label a large number of tweets (226k) using ternary classes. They found that TextBlob labelling generated normalized scales of the sentiment labels compared to the Afinn labelling and obtained 81\% accuracy in the LR classification method. They found the TextBlob labelling generated normalized scales of the sentiment labels compared to the Afinn labelling and obtained 81\% accuracy in the LR classification method. 
Saad et al. (2021) used Afinn, TextBlob, and Vader to assign sentiment labels to a drug review dataset, and they obtained 96\% accuracy with TextBlob; similarly, \cite{hasan2018machine} and \cite{wadera2020sentiment} also got success with TextBlob labelling on Twitter datasets using some of the traditional machine learning algorithms like Support Vector Machine (SVM), Naïve Bayes (NB), Random Forest Tree (RFT), Decision Trees (DTC), and K-Nearest Neighbours (KNN). There is no one automatic labelling technique that does best consistently in the previous work, and all the evaluations used either human labelling or automatic labelling, but they did not use both labelling strategies and did not compare them.

State-of-art deep learning algorithms have been used to predict the sentiment of text that is shared on different social media in several studies \cite{poria2015deep,lei2016rating}. Liao et al. (2017) proposed a one-layer Convolution Neural Network (CNN) to analyse the sentiment. Similarly, Long Short Term Memory (LSTM), Bidirectional (BiLSTM) models are also used in natural language processing \cite{xu2019sentiment}.  However, all the above research used human labelling in their state-of-the-art deep learning algorithms, but they did not use automatic labelling.



\begin{figure*}[!h]
	\centering
	{\includegraphics[width=12cm,height=6cm]{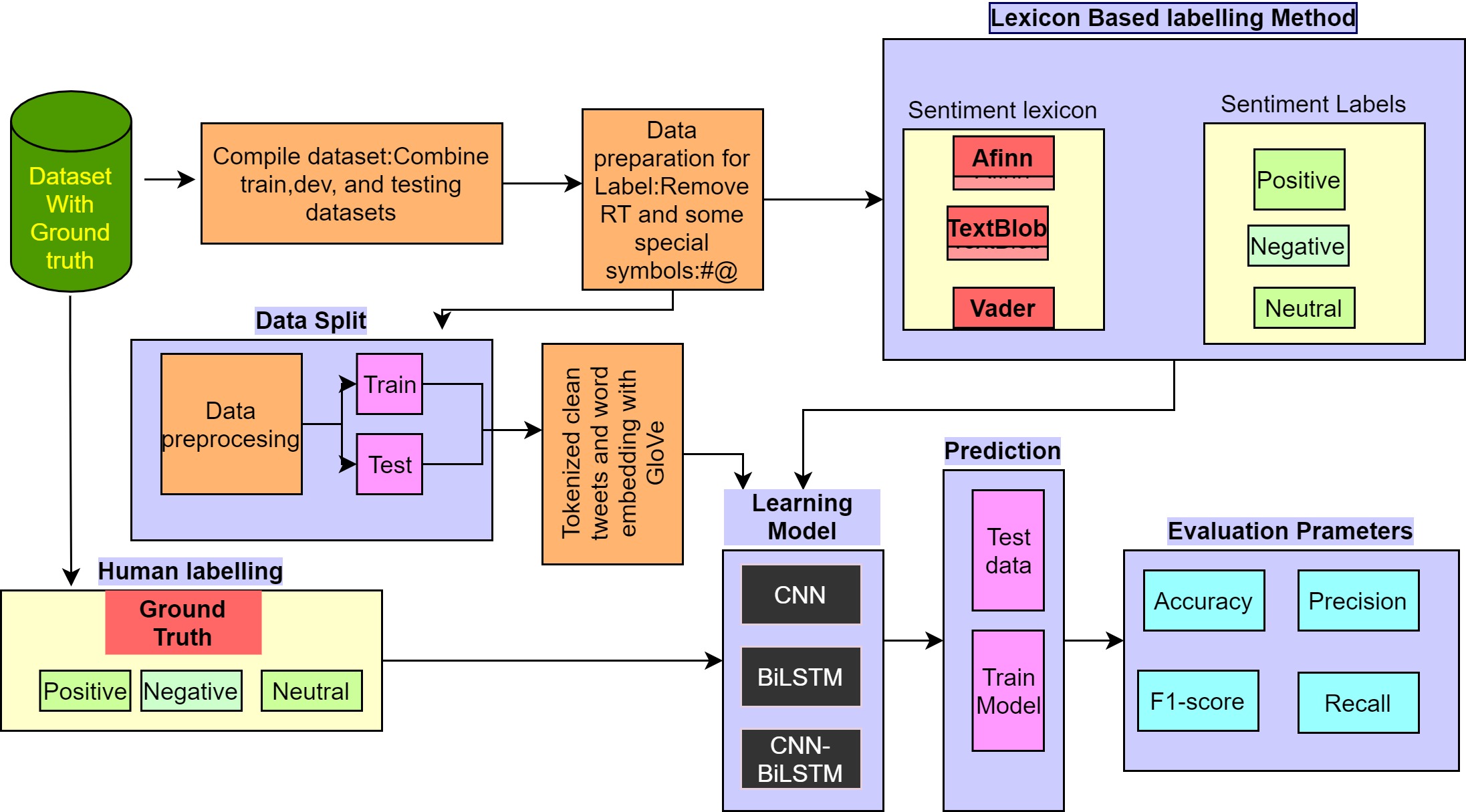}}
	\caption{A summary of the experiments performed.}
	\label{fig:example1}
\end{figure*}
\section{\uppercase{Methodology}}
\label{sec:methodology}
We have used two Twitter datasets from the SemEval-$13$ and SemEval-$16$ competition which contains human labels (HL). We did not consider the remaining datasets from this competition because they had either imbalanced human labels or had a small number of tweets. The details of the datasets are given in the paper \cite{deriu2016swisscheese} and \cite{yoon2017multi}. SemEval-2013 Task 4 and SemEval-2016 Task-4 are named as DS-1 and DS-2 respectively in this study. The original datasets are divided into train, dev, and test. We combine the training, development, and testing portions of each dataset to create a single whole dataset. The first experiment uses the existing SemEval labels (human labels) for both training and testing. The second experiment uses, in turn, the three automatic labelling methods (TextBlob, Vader, and Afinn) for both the training and testing datasets. The third experiment uses the existing SemEval labels (human labels) for testing, but the training data labels are generated using, in turn, the three automatic labelling methods (TextBlob, Vader, and Afinn). In all cases, 80\% of each dataset is used for training and the remaining 20\% of the dataset is used for testing. For each experiment, CNN, BiSLTM, and CNN-BiSLTM deep learning algorithms are applied to the cleaned and pre-processed training data and the models are tested on the test data.
Figure 1 illustrates the details of the methodology which is explained in the following steps.

\subsection{Data Labelling}
In experiment 2 for both training and testing datasets and in experiment 3 for only the training dataset, we replace the human labels with automatically generated labels. Prior to assigning the automatic labels, we remove the username and some special symbols such as\ @, \#, \$, and RT from all tweets in the datasets. We have used the Natural Language Toolkit (NLTK) library from Python to label the tweet’s sentiment as Positive, Negative or Neutral using the following three methods:

 \begin{table*}[h]
	\vspace{-0.2cm}
	\hspace*{1.5cm}\textit{$\backslash$end\{table*\}}\\
	\caption{Summary statistics of the total number of tweets from the SemEval competitions Tweet-2013 (DS-1) and Tweet-2016 (DS-2), and the percentage of Positive (Pos.), Negative (Neg.), and Neutral (Neut.) sentiment labels from human labelling and automatic labelling approaches (TextBlob, Vader, Afinn).}
	\label{tab:example1} \centering
	
	\begin{center}
		\begin{tabular}
			{ |p{.88cm}|p{.85cm}|p{.65cm}|p{.65cm}|p{.55cm}|p{.65cm}|p{.65cm}|p{.65cm}|p{.65cm}|p{.65cm}|p{.65cm}|p{.65cm}|p{.65cm}|p{.65cm}|}

			\hline
			Dataset&Total Tweets &\multicolumn{3}{c|}{Human Labelling(\%)} &\multicolumn{9}{c|}{Automatic Labelling(\%)} \\
			\hline
			& & & & &\multicolumn{3}{c|}{TextBlob} &\multicolumn{3}{c|}{Vader}& \multicolumn{3}{c|}{Afinn} \\
			\hline
			&  &Pos&Neg&Neut&Pos&Neg&Neut&Pos&Neg&Neut&Pos&Neg&Neut\\
			\hline
			DS-1&14885&38.23&15.84&45.94&48.66&19.55&31.41&51.55&18.2&29.99&45.76&17.96&36.01\\ 
			DS-2&28631&38.41&15.67&45.93&46.34&20.28&32.4&47.06&24.48&27.48&42.1&22.87&34.05 \\ 
			\hline
		\end{tabular}
		
	\end{center}
	\hspace*{1.5cm}
\end{table*}  

TextBlob:  In this study, we have used the polarity of sentiment for textual data among the many properties available as part of TextBlob in the Python library. TextBlob returns a polarity value within the range [-1.0, 1.0] where ‘-1’ indicates a very Negative polarity, ‘0’ a Neutral polarity and ‘+1’ is a very Positive polarity.  

Vader: VADER stands for ‘Valence Aware Dictionary and sEntiment Reasoner’ \cite{hutto2014vader}. It is a rule-based sentiment analysis tool, which generates the scores of sentiments by the intensity of lexical features and semantic meanings of the text. Vader returns four components with associated intensity scores. For example, \{neg: 0.106, neu: 0.769, pos: 0.125, compound: 0.1027\}, where ‘neg’, ‘neu’, ‘pos’ indicate the Negative, Neutral, and Positive scores respectively and the compound score is the normalized score of the summation of the valence scores computed based on heuristic and lexicon sentiments. In this study, we assign ‘Positive’ label when the compound score is greater than 0.05, ‘Negative’ label when the compound score is less than -0.05 and otherwise, we assign ‘Neutral’ label.

Afinn: Afinn is a simple and popular lexicon developed by Finn Arup Nielsen \cite{nielsen2011new}. Afinn returns the score of a word between [-5, 5]. Here we assign a ‘Positive’ label when the score is greater than ‘0’, a ‘Negative’ label when the score is less than ‘0’ and otherwise we assign a ‘Neutral’ label. 

To illustrate the automatic labelling techniques we take as an example from DS-2: “install the newest version and you may change your mind!”; the sentiment label is assigned as ‘Neutral’, ‘Positive’, and ‘Positive’ using the TextBlob, Vader and Afinn methods respectively. It is noted that the existing human (ground truth) label of the same text is also ‘Positive’.

Table 1 presents the two datasets with the original name, the newly assigned name in our experiments, and the total number of tweets associated with each dataset. It also shows the summary of the percentage of sentiment labels per dataset with the automatic labelling approaches in comparison to the human labels. Positive labels are remarkably high when using automatic labelling techniques. On the other hand, Neutral labels are notably high in the human labelled datasets.

\subsection{Data Preprocessing for Models }
Data pre-processing and cleaning is an important step when applying deep learning algorithms to the data. We remove any unwanted words using a list of stop words from the NLTK (Natural Language Toolkit) library in Python, without removing stop words like ‘our’, ‘ours’, ‘ourselves’, ‘these’, ‘those’ etc. We removed all digits in the tweets and some unwanted words which did not make any sense like ‘amp’, ‘tiktok’, ‘th’, etc., and inappropriate words. The cleaned tweets are then tokenized after being split into training and testing sets.  We have used GloVe of 300 dimensions for word embedding. GloVe, which stands for Global Vectors, is an unsupervised learning algorithm for achieving vector representations of words. Large quantities of text are used for training and converting into low dimensional and dense word formations in an unsupervised fashion by an embedding process \cite{pennington2014glove}. One hot encoding is a highly popular technique to deal with categorical data, which creates a binary column for each category and returns a sparse matrix or dense array based on the sparse parameter. We have labelled the sentiments of tweets in three categories and used the one-hot encoding method. For this, three new columns are created as ‘Positive’, ‘Negative’, and ‘Neutral’, and these dummy variables are then filled up with Zeros (meaning False) and ones (meaning True).
\begin{figure}[!h]
	\centering
	\label{key}
	{\includegraphics[width=5cm,height=8cm]{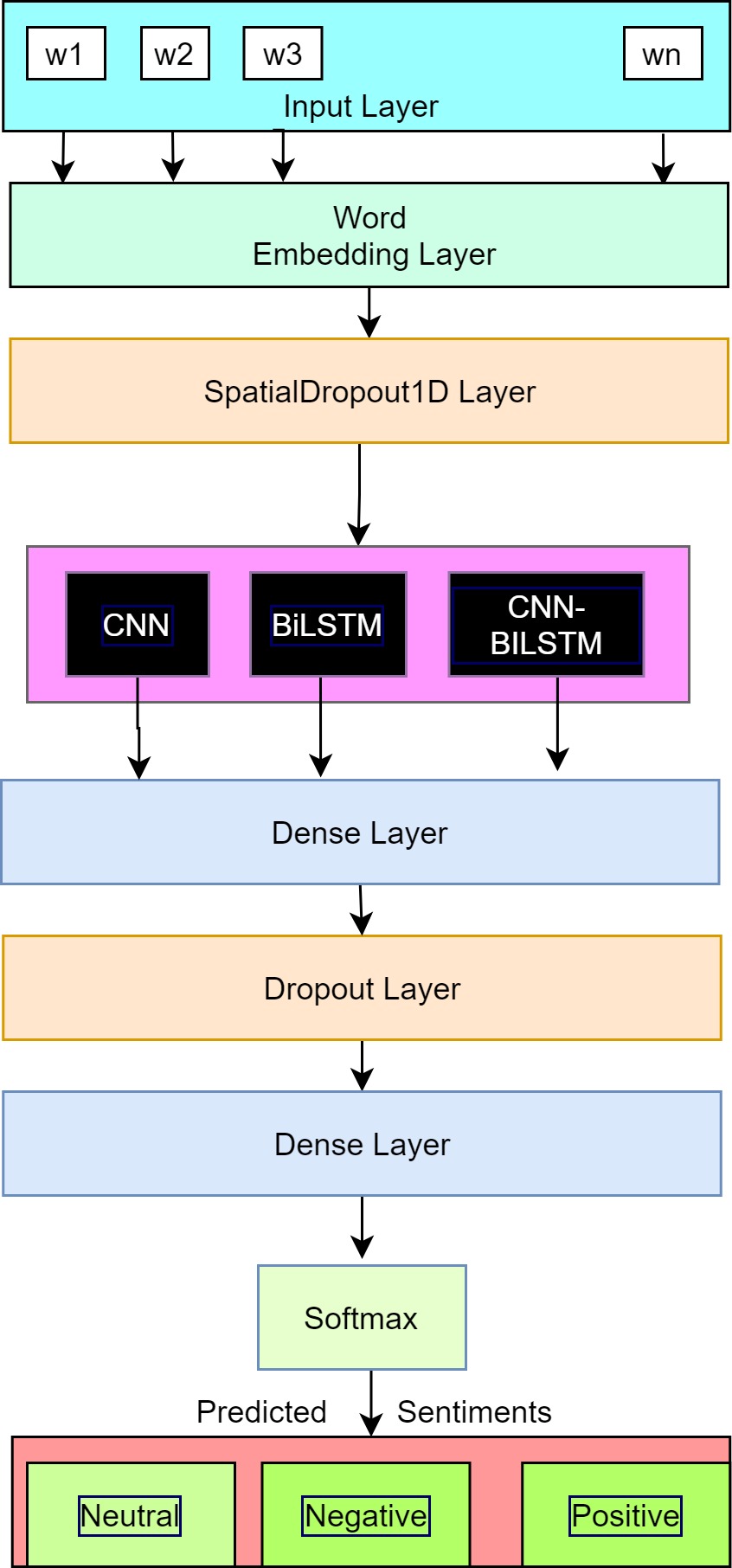}}
	
	
	\caption{Structure of the deep learning models.}
	\label{fig:example1}
\hspace*{1.5cm}	
\end{figure}
\subsection{Deep Learning Models}
This section describes the method of using deep learning models. CNN (Convolution Neural Network), BiLSTM (Bidirectional Long Short-Term Memory), and a combined model of CNN-BiLSTM are used to find the sentiments of tweets. The model consists of some common layers: an embedded layer, a SpatialDropout1D layer, a deep learning layer, Dense layer, Relu dense layer, Dense, and finally a Softmax layer to predict sentiments. Three deep learning models: CNN, BiLSTM, and CNN-BiLSTM are used separately in the deep learning layer, which is in between the SpatialDropout1D layer and the Dense layer of the common layers. The architecture of the models is shown in Figure 2. A SpatialDropout1D layer takes a word embedding matrix of an input sentence; It helps to prevent pixels from co-adapting with their neighbors across feature maps, which promotes independence between feature maps. The output is fed to different deep learning networks. Each output of the three deep learning models is fed separately into a dense layer with a Relu activation function, and then a final dense layer with a softmax activation function predicts the probabilities of ternary sentiment labels.

CNN: A CNN is a feed-forward neural network containing an input layer, hidden layer and an output layer, which is capable of capturing all local features. It computes the most important features from the output of the CNN \cite{liao2017cnn} with a Rectified Linear Unit (ReLU) activation, and Global max pooling layer.

BiLSTM: Long Short-Term Memory (LSTM) is a Recurrent Neural Network (RNN) with three gates: input gate, output gate, and forgot gate \cite{xu2019sentiment}. It has a forward and backward layer. BiLSTM is capable of remembering future and past information from input sequences and processing the information in both directions. 

CNN-BiLSTM: A CNN model is combined with a BiLSTM \cite{xu2019sentiment} model. CNN-BiLSTM includes all prominent level features and long-term dependencies in both directions \cite{chaturvedi2018distinguishing} of input datasets for all labelling strategies in the aforementioned three experiments. 
\subsection{Experiment Setup}
Each deep learning model takes inputs from the SpatialDropout1D layer which is embedded with sequence input. The maximum input sequence length is 30. The CNN layer uses 64 filters with kernel size 5. The BiLSTM layer is used with 64 hidden units. A fully connected layer also is used with 64 hidden units. To avoid overfitting and underfitting, CNN uses two dropout levels and BiLSTM and CNN-BiLSTM models employ four dropout levels. To get good results for different algorithms, we chose varied drop out rates. For example, in the first experiment, the drop out rates in the CNN model are (0.2 and 0.2), however in the second and third experiments, the drop out rates are (0.4, 0.5) and (0.3, 0.5), respectively. In the first, second, and third experiments, the dropout rates for BiLSTM and CNN-BiLSTM are (0.3,0.3,0.3,0.4), (0.2,0.2,0.2,0.5), and (0.3,0.3,0.3,0.5), respectively. We have set a learning rate of 0.0001 for the first experiment (human labelling). A learning rate of 0.001 is considered for the second and third experiments. Categorical cross-entropy is used as the loss function and categorical accuracy is used for the accuracy metric. We set the batch size to be 100 and run for 10 epochs to average the metrics for accuracy, precision, F1 scores and recall.

\begin{table*}[h]
	\caption{Experiment 1: Comparison of Accuracy, Precision, F1-score and Recall with human labels using a CNN, BiLSTM, and CNN-BiLSTM Model.}
	\label{tab:example1} \centering
	\begin{center}
		\begin{tabular}
			{ |p{3cm}|p{1cm}|p{1cm}|p{1cm}|p{1cm}|p{1cm}|p{1cm}|p{1cm}|p{1cm}|  }
			
			\hline
			{Model} &\multicolumn{4}{c|}{DS-1(\%)} &\multicolumn{4}{c|}{DS-2(\%)} \\
			\hline
			&Acc&Pre&F1&Rec&Acc&Pre&F1&Rec \\
			\hline
			CNN+HL&	60.36&	65.50&	55.19&	52.08&	54.97&{\textbf{66.92}}&	56.39&	50.48 \\
			\hline
			BiLSTM+HL&{\textbf{61.55}}&	{\textbf{71.95}}&	{\textbf{58.25}}&	{\textbf{54.25}}&	{\textbf{60.54}}&	66.79&	{\textbf{61.99}}&	{\textbf{57.99}} \\
			\hline
			CNN-BiLSTM+HL&	60.62&	63.82&	55.97&	53.99&	57.06&	66.14&	57.74&	52.93 \\
			\hline
		\end{tabular}
	\end{center}
\end{table*} 

\begin{table*}[h]
	\caption{Experiment 2: Comparison of Accuracy, Precision, F1-Score and Recall with automatic labelling used for both testing and training datasets using CNN, BiLSTM, and CNN-BiLSTM Models.} 
	\label{tab:example1}
	\begin{center}
		\begin{tabular}
			{ |p{3cm}|p{1cm}|p{1cm}|p{1cm}|p{1cm}|p{1cm}|p{1cm}|p{1cm}|p{1cm}|  }
			
			\hline
			{Model} &\multicolumn{4}{c|}{DS-1(\%)} &\multicolumn{4}{c|}{DS-2(\%)} \\
			\hline
			&Acc&Pre&F1&Rec&Acc&Pre&F1&Rec \\
			\hline
			CNN+Af	&74.26&	76.51&	72.24&	69.37&	74.28&	77.48&	74.39&	71.67 \\
			BiLSTM+Af&{\textbf{80.17}}	&{\textbf{81.63}}&{\textbf{77.54}}	&{\textbf{74.92}}	&{\textbf{80.05}}	&{\textbf{84.06}}	&{\textbf{80.70}}&{\textbf{77.80}}	 \\
			CNN-BiLSTM+Af&	75.27&	78.11&	73.04&	69.79&	76.9&	81.16&	77.9&	74.93 \\
			CNN+Tb&	63.17&	73.70&	61.37&	55.60&	69.27&	76.09&	69.69	&65.8 \\
			BiLSTM+Tb&	73.41&	77.38&	72.06&	68.75&	79.55&	82.67&	78.77&	76.44 \\
			CNN-BiLSTM+Tb&	66.95&	72.77&	65.00&	61.15&	75.03&	78.46&	74.5&	71.92 \\
			CNN+Vd&	68.22&	72.8&	66.17&	62.37&	68.19&	76.09&	71.04&	67.03 \\
			BiLSTM+Vd&	73.75&	75.99&	70.64&	68.53&	76.79&	79.29&	76.63&	75.0 6\\
			CNN-BiLSTM+Vd&	71.12&	73.46&	66.96&	63.81&	72.4&	76.58&	73.56&	70.89 \\
			\hline
		\end{tabular}
	\end{center}
\end{table*} 

\begin{table*}[h]
	\caption{Experiment 3: Comparison of Accuracy, Precision, F1-score and Recall with automatic labelling techniques (training dataset) and human labelling for the test set using CNN, BiLSTM, and CNN-BiLSTM Models.} 
	\label{tab:example1}
	\begin{center}
		\begin{tabular}
			{ |p{3cm}|p{1cm}|p{1cm}|p{1cm}|p{1cm}|p{1cm}|p{1cm}|p{1cm}|p{1cm}|  }
			
			\hline
			{Model} &\multicolumn{4}{c|}{DS-1(\%)} &\multicolumn{4}{c|}{DS-2(\%)} \\
			\hline
			&Acc&Pre&F1&Rec&Acc&Pre&F1&Rec \\
			\hline
			CNN+Af&	57.49&	60.03&	54.91&	51.97&	46.57&	53.18&	49.08&	46.68\\
			BiLSTM+Af&	61.61&	63.78	&58.03&	55.02&	46.46& {\textbf{54.76}} &{\textbf{50.07}} &	46.69\\
			CNN-BiLSTM+Af& {\textbf{62.74}} &{\textbf{63.74}} &{\textbf{58.08}} &{\textbf{55.76}} &{\textbf{53.20}} &53.21 &	47.98&	45.67 \\
			
			CNN+Tb	&48.71&	49.74&	46.07	&45.75&	48.47&	46.27&	45.07&	44.95 \\
			BiLSTM+Tb&	55.91&	63.79	&52.67&	48.52&	49.47&	50.27&	45.07&	44.95\\
			CNN-BiLSTM+Tb&	55.48&	54.50&	48.66&	45.35&	45.57&	46.85&	44.25&	42.79\\
			CNN+Vd	&54.24&	56.64&	51.78&	50.49&	47.91&	50.82&	48.96&	49.45\\
			BiLSTM+Vd&	55.32	&60.29	&53.19&	50.65	&48.24&	51.08&	48.6&{\textbf{50.33}}\\
			CNN-BiLSTM+Vd&	57.82&	62.36&	54.28&	51.53	&46.31&	51.97	&48.04&	46.31\\
			
			\hline	
		\end{tabular}
	\end{center}
\end{table*} 

\section{\uppercase{Experiment Results and Discussion}}
\label{sec:experiment result and discussion}
The performance of the models is measured by using four possible outcomes: True Positives (TP), True Negatives (TN), False Positive (FP), and False Negative (FN). The accuracy of the result is computed using those outcomes, but accuracy can mislead in imbalanced datasets. This is why, we have computed precision, recall, and F1-score. Table 2 shows the results of the first experiment, which used three deep learning models with human labeling for the two datasets. The results are evaluated in macro averaged performance across the Positive, Negative, and Neutral classes. The BiLSTM model obtains the highest accuracy, precision, F1-score and recall for the two datasets. The best results are highlighted in bold in the result Tables. The accuracy of the BiLSTM model is 1.19\% higher than the CNN model and 0.93\% higher than the CNN-BiLSTM model for DS-1. In the same way, the accuracy of the BiLSTM model is 5.57 and 3.48\% higher than the accuracy of the CNN and CNN-BiLSTM model respectively for DS-2. The F1-score is also higher with the BiSLTM model than the other two models.

The results of the second experiment are shown in Table 3 using the three automatic labelling strategies and the three deep learning algorithms. The results show higher accuracy, precision, recall, and F1-scores for the two datasets with automatic labelling than human labelling. Additionally, Afinn labelling obtains the highest accuracy using the BiLSTM model for the two datasets compared to the other models with TextBlob and Vader labelling techniques. The balance of sentiment labels are better in Afinn labelling than the other labelling techniques, and this could explain why Afinn does better than the other labelling strategies. On the other hand, the CNN model could not learn all dependencies of whole sentences due to the limitation of filter lengths \cite{camgozlu2020analysis}. As a result, CNN and CNN-BiLSTM models were not able to learn all features from sentences or whole datasets. For this reason, BiLSTM outperformed the other models with F1 scores of 77.54\% and 80.7\% for DS-1 and DS-2 respectively.

In the third experiment, we have observed the results with 80\% of the training data with automatic labelling and 20\% of test data with human labelling. Table 4 displays the results of the third experiment for the two datasets. The aim is to create a strenuous test set to assess if the model can perform well when tested with human labelling after being trained using automatic labelling. The accuracy, precision, F1-score, and recall values for the CNN-BiLSTM model were the best. These results are 2.12\%, 2.29\%, and 1.77\% higher in accuracy, F1-score, and recall values, respectively, than the results of the first experiment for DS-1, though the precision value is 0.08\% lower. Similarly, the BiLSTM model obtained the highest precision, F1-score, and recall, but the CNN-BiLSTM model obtained the highest accuracy for DS-2. The results are better in Experiment 3 than in Experiment 1. The model did best with Afinn labelling. 

As can be seen with previous work \cite{chakraborty2020sentiment,saad2021determining,deepa2019sentiment,zahoor2020twitter} there is no one automatic labelling technique that does best consistently. In our work, the Afinn labelling technique almost always produces the best results, across different models and experiments (Except in one case, the best recall value is obtained for DS2 in the Vader labeling technique). This labelling technique uses only sentiment of words (unlike Vader) and may be better suited to Tweet data for this reason. 

When considering the performance of the deep learning models, the BiLSTM model does best in most situations in Experiment 1 and Experiment 2. In the third experiment, the CNN-BiLSTM model does best with DS-1 but not with DS-2.

\begin{figure}[!h]
	\centering
	
	{\includegraphics[width=7cm,height=4cm]{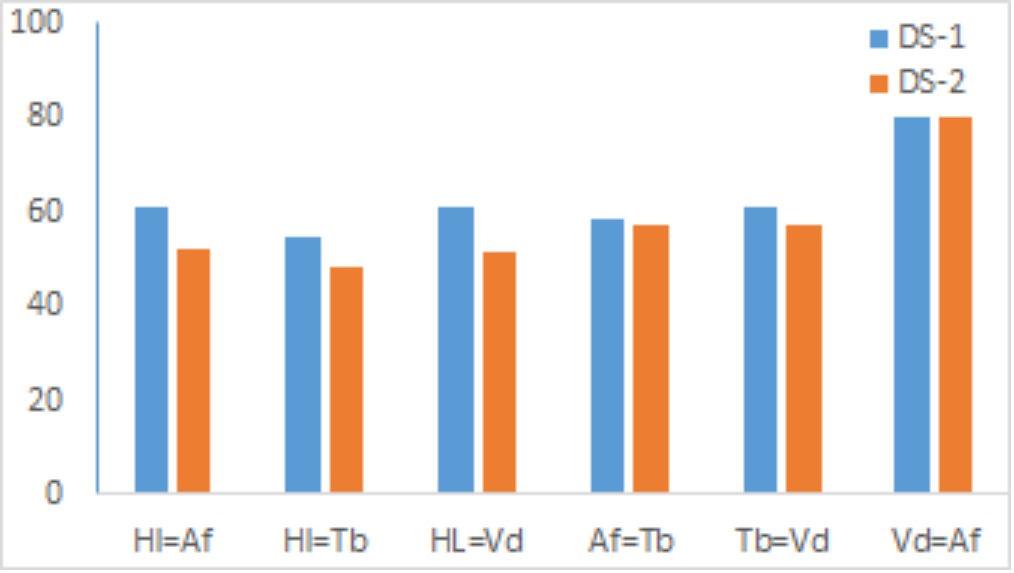}}
	
	
	\caption{Percentages of tweets that are labelled with the same sentiment label by the human labellers and the automatic techniques: Afinn, TextBlob and Vader.}
	\label{fig:example1}
	\hspace*{1.5cm}
\end{figure}

In Figure 3, ‘HL=Af’ represents the percentage of sentiment labels that are the same across the human labelling and Afinn labelling approaches. Similarly, ‘HL=Tb’ and’ ‘HL=Vd’ represent the percentage of labels that are the same across the human labelling approach and the TextBlob labelling and Vader labelling respectively. The three automatic labelling approaches: Afinn, TextBlob and Vader obtain an average of 59\% and 50.55\% of equal sentiment labels with the human labelling for DS-1 and DS-2 respectively. On the other hand, there is a much higher overlap of equal sentiment labels between Vader and Afinn (80\%).
We can also see in Figure 3 across the different analyses and experiments that there is inconsistency with the labels given by human labellers and by the automatic labelling techniques with the most agreement existing between Vader and Afinn labelling techniques.  In Figure 3, there is, at best, a difference of 41\% (and sometimes more) between the human labels and automatic labels. This explains why the performance results in Experiment 3 are often poorer than those reported in Experiment 1 and 2. The learning task is much more difficult given that the labels for the training data and the labels for the test data are generated in different ways.

The best results are shown in Experiment 2 with the automatic labelling approaches being used for both the training and test sets. We argue that because the same lexicon and rules have been used to label both the training and test set it is easier for a machine learning approach to learn these rules and make more accurate predictions.  We suggest that this is not the case with human labelling and hence the machine learning models find it more difficult to do as well in Experiment 1.
\section{\uppercase{Conclusion}}
\label{sec:conclusion}
These experiments highlight the care that should be taken in the use of both automatic and human labelling approaches. After evaluating both approaches, we compared the performance of the three leading automatic ternary labelling strategies to establish automatic labelling as a feasible alternative to human labelling. This automated approach to sentiment labelling would yield extensive benefits by eliminating the effort and cost of human labelling. Our results have identified the Afinn approach as having the highest level of labelling accuracy with both datasets. We suggest that because both the training and test sets were labelled with the same lexicon and rules, a machine learning technique would find it easier to learn these rules and produce more accurate predictions. These experiments motivate the need for further investigations into the differences between the different automatic approaches as well as the differences between human labelling and the automatic labelling approaches. However, the limitations of this study include the potential mislabelling of data using automated approaches. We have evaluated the experiments on a relatively small size of the dataset with only three labelling approaches, and this proposed methodology might not be suitable where the training dataset is smaller than the testing dataset. In the future, we want to continue this experiment for other large text-classification Twitter datasets, for example, Covid19 or Covid-Vaccine related datasets. We will also explore and extend this study to analyse the sentiment using other deep learning models with different embedding methods, for example, BERT model with word2vec and fast-text in the future. We hope these studies would be helpful to analyse sentiment using automatic labelling on large datasets when we want to save time and cost for generating ground truth by humans.
\section*{\uppercase{Acknowledgements}}
This work was supported by the College of Engineering, National University of Ireland, Galway.

\bibliographystyle{apalike}
{\small
\bibliography{data}}

\end{document}